\documentclass[10pt, a4paper]{article}

\usepackage[final]{lrec2026}

\usepackage{soul}
\usepackage{amsmath}
\usepackage{booktabs}
\usepackage{makecell}
\usepackage{comment}

\usepackage{adjustbox}
\usepackage[utf8]{inputenc}
\usepackage{multirow} 

\usepackage{graphicx}
\usepackage{placeins}

\usepackage{listings}
\usepackage{dblfloatfix}

\usepackage{colortbl}
\definecolor{grey}{rgb}{.8, .8, .8}

\title{Low-Resource Dialect Adaptation of Large Language Models: \\ A French Dialect Case-Study}

\name{
  Eeham Khan\textsuperscript{1},
  Firas Saidani\textsuperscript{2}, 
  Owen Van Esbroeck\textsuperscript{1},
  Richard Khoury\textsuperscript{2},
  Leila Kosseim\textsuperscript{1}
}

\address{
  \textsuperscript{1}\,Computational Linguistics at Concordia (CLaC) Laboratory\\
  Department of Computer Science and Software Engineering\\
  Concordia University, Montr\'eal, Qu\'ebec, Canada\\
  \texttt{\{eeham.khan, leila.kosseim\}@concordia.ca, owen.vanesbroeck@mail.concordia.ca}\\[0.7em]
  \textsuperscript{2}\,Group for Research in Artificial Intelligence of Laval University (GRAIL)\\
  Department of Computer Science and Software Engineering\\
  Universit\'e Laval, Qu\'ebec, Canada\\
  \texttt{\{richard.khoury, firas-mustapha.saidani.1\}@ulaval.ca}
}

\abstract{Despite the widespread adoption of Large Language Models (LLMs), their strongest capabilities remain largely confined to a small number of high-resource languages for which there is abundant training data. Recently, continual pre-training (CPT) has emerged as a means to fine-tune these models to low-resource regional dialects. In this paper, we study the use of CPT for dialect learning under tight data and compute budgets. Using low-rank adaptation (LoRA) and compute-efficient continual pre-training, we adapt three LLMs to the Québec French dialect using a  very small dataset and benchmark them on the COLE suite. Our experiments demonstrate an improvement on the minority dialect benchmarks with minimal regression on the prestige language benchmarks with around 1\% of model parameters updated. Analysis of the results demonstrate that gains are highly contingent on corpus composition. These findings indicate that CPT with parameter-efficient fine-tuning (PEFT) can narrow the dialect gap by providing cost-effective and sustainable language resource creation, expanding high-quality LLM access to minority linguistic communities. To support reproducibility and broaden access, we release the first Québec French LLMs on Hugging Face. 
\\ 
}

\begin{document}

\maketitleabstract

\section{Introduction}
\label{sec:intro}
In recent years, Large Language Models (LLMs) have emerged as powerful and versatile tools for many applications, driving progress in tasks such as text summarization, text and code generation, and open-domain dialogue systems. These LLMs are pretrained on extensive corpora to maximize general-purpose performance across a wide variety of downstream tasks. 

Despite their success, most widely-used LLMs are trained on predominantly English datasets. When other languages are included in multilingual training datasets, they are collected from the high-resource prestige dialect of the language and have little coverage of low-resource regional dialects. As a result, they often struggle with local vocabulary, orthographic variants, idiomatic expressions, and code-switching phenomena that are commonly found in regional dialects. This issue, coined the \textit{dialect gap}~\cite{kantharuban2023quantifying}, limits their effectiveness for millions of users of these minority language varieties, thereby perpetuating inequities in access to artificial intelligence technologies.

One possible approach to addressing the dialect gap is to train LLMs using regional dialectal data. However, full model training is prohibitively expensive, while conventional fine-tuning approaches can be inefficient or lead to overfitting on relatively small dialectal datasets. Continual Pre-training (CPT) on unlabeled data offers a practical compromise: by exposing models to large volumes of dialect-specific text, CPT enhances regional linguistic coverage without fully discarding the general knowledge encoded during initial pre-training~\citep{gururangan-etal-2020-dont,sarkar2022parameter,lee2019biobert}.

In this paper, we explore the ability and limitations of CPT to adapt LLMs to a regional dialect using a small amount of unlabeled regional texts. As a case study, we consider specifically the case of the Qu\'{e}bec regional dialect of French, also called Qu\'{e}b\'{e}cois. 
To make adaptation feasible with less compute resources, we employ low-rank adaptation (LoRA) and gradient checkpointing as parameter-efficient strategies. The adapted models are then benchmarked on a subset of the COLE suite of French tasks~\cite{beauchemin2025cole}, including both Qu\'{e}bec French and prestige French benchmarks, to assess their ability to balance dialectal adaptation while retaining general competence in French.


Our key contributions in this paper are:
\begin{enumerate}
    \item 
    We demonstrate a compute-efficient CPT pipeline (LoRA + gradient checkpointing) that updates $\leq\!1\%$ of parameters while sustaining performance on modest hardware (see Section~\ref{sec:method}). We show that it is possible to adapt an LLM to a regional dialect using only a very small corpus of 86M tokens in our experiments, which is significantly below typical CPT data budgets.
    
    \item 
    Using this methodology, we train and release the first open-weight LLMs adapted specifically to Qu\'ebec French, and evaluate them on a subset of 
    the COLE benchmark suite. These tests highlight the capacity and limitations of CPT dialectal training (Section~\ref{sec:analysis}).

    \item We provide complete training configurations, data-processing scripts, and evaluation pipelines for direct transfer to other dialects and low-resource varieties on GitHub\footnote{\label{github}\url{https://github.com/CLaC-Lab/QuebecLLM-CPT}}. We also release our Qu\'ebec French LLMs on HuggingFace\footnote{\label{HuggingFace}\url{https://huggingface.co/QuebecLLM}}.

\end{enumerate}

\section{Background and Related Work}
\label{sec:related}

\subsection{Continual Pre-training}
Continual Pre-training continues training a model for its original self-supervised objective (e.g., masked or causal language modeling) using an out-of-domain corpus. The objective is to extend the model's capabilities using additional, unlabeled datasets. CPT differs from supervised fine-tuning, which uses smaller task-specific labeled datasets. \\

Although CPT is related to continual learning (CL), it is more appropriate for practical specialization rather than lifelong multi-task learning. CL methods such as regularization~\citep{Kirkpatrick_2017,aljundi2018memoryawaresynapseslearning}, replay~\citep{lopezpaz2017gem,chaudhry2019tinyepisodicmemoriescontinual}, and gradient orthogonalization~\citep{farajtabar2020orthogonal} mitigate catastrophic forgetting across many tasks. On the other hand, CPT typically adapts to a \emph{single} new distribution without explicit replay or regularizers, retaining the model's original capability through careful optimization and small learning rates~\citep{gururangan-etal-2020-dont}.\\

While full-parameter fine-tuning is often prohibitive for LLMs due to compute time and cost, low-rank adaptation (LoRA)~\citep{hu2022lora} allows updates to only a small fraction of model parameters while preserving pretrained weights. Recent advances combine LoRA with quantization-aware training (e.g., L4Q~\citep{jeon2025l4q}) to further reduce costs. Other parameter-efficient fine-tuning (PEFT) techniques such as adapters~\citep{houlsby2019parameter,pfeiffer-etal-2021-adapterfusion}, prefix/prompt tuning~\citep{li2021prefixtuning,lester-etal-2021-power}, BitFit~\citep{ben-zaken-etal-2022-bitfit}, and learned attention/FFN scaling~\citep{liu2022fewshot} reduce memory and computation costs while achieving competitive performances. For resource-constrained continual pre-training, LoRA and related methods can be combined with mixed precision, gradient checkpointing, and quantization to keep adaptation feasible~\citep{dettmers2023qlora}. 

\subsection{Dialect Adaptation}

Adapting pretrained LLMs to new linguistic varieties can be framed as \emph{domain-adaptive pre-training} (DAPT) or \emph{language-adaptive pre-training} (LAPT), where models are exposed to additional unlabeled target-distribution text~\citep{gururangan-etal-2020-dont,howard2018universal,chronopoulou-etal-2019-embarrassingly}. This improves coverage of distribution-specific lexical items, morphology, and style while preserving general knowledge of the base model. Prior work has applied this to new domains (e.g. biomedical or legal) \cite{lee2019biobert, chalkidis-etal-2020-legal} and low-resource languages~\citep{beltagy2019scibert,koto2025sherkalachat,mansha2025resourceefficient}. 

CPT has become a dominant approach for adapting LLMs to low-resource languages and regional dialects~\cite{elhady2025emergent}. 

Such systems have been developed for a variety of languages and dialects, trained on datasets as small as 350M tokens.
Recent work also shows that mixing English with the target language during CPT can be critical for preserving downstream abilities~\cite{elhady2025emergent}.

\begin{table}[t]
\centering
\scriptsize
\setlength{\tabcolsep}{3pt}
\renewcommand{\arraystretch}{1.05}
\resizebox{\columnwidth}{!}{%
\begin{tabular}{l r c}
\toprule
\textbf{Language} & \textbf{Tokens} & \textbf{Reference} \\
\midrule
Qu\'{e}b\'{e}cois & 0.085B & Ours \\
Galician (Carballo) & 0.340B  &~\cite{caraballo2024} \\
Catalan (CataLlama) & 0.445B  &~\cite{cataLlama2024} \\
Sorbian & 1.2B  &~\cite{turtunlp2025sorbian} \\
Basque (Latxa) & 4.2B  &~\cite{agerri2024latxa} \\
Kazakh (Sherkala-Chat) & 45B & \citep{koto2025sherkalachat} \\
German (LeoLM) & 65B  &~\cite{leolm2023} \\
Hindi (Nemotron-Mini) & 400B &~\cite{dabre2025nemotron} \\
Vietnamese (VinaLlama) & 500B  &~\cite{vinaLlama2023} \\
Arabic (ALLaM) & 600B to 1200B  &~\cite{allam2024} \\
\bottomrule
\end{tabular}}
\caption{Regional languages for which LLMs have been trained using CPT, along with the number of training tokens.}
\label{tab:tokens}
\end{table}

French LLMs such as CamemBERT and FlauBERT established strong monolingual baselines~\citep{martin-etal-2020-camembert,le-etal-2020-flaubert}, while multilingual encoders (e.g., XLM-RoBERTa) remain competitive for French~\citep{conneau-etal-2020-unsupervised}. However, these are trained on the prestige dialect of French used in France. 
The Qu\'{e}bec dialect of French, or Qu\'{e}b\'{e}cois, differs from the prestige dialect in orthography, phonology-to-orthography conventions, anglicisms, idioms, and code-switching patterns. In addition, Qu\'{e}b\'{e}cois resources are scarce, compared to prestige French or to other languages listed in Table~\ref{tab:tokens}. These characteristics make CPT a viable option for dialectal adaptation. 

\section{Datasets}
\label{sec:data}

\subsection{Data sources}
\label{sec:data-sources}
To adapt models to  Québécois, we collected a corpus of documents from a variety of Qu\'{e}bec French sources spanning news, blogs, transcribed speech, and social media. All copyrighted materials were obtained and used with the explicit permission of the rights holders. 

\begin{table}[t]
\centering
\footnotesize
\setlength{\tabcolsep}{4pt}
\renewcommand{\arraystretch}{1.08}
\resizebox{\columnwidth}{!}{%
\begin{tabular}{r l l c r l}
\toprule
& \textbf{Source} & \textbf{Type} & \textbf{Years} & \textbf{Size} & \textbf{License}\\
\midrule
1.& BEQ ebooks                     & Books        & 1800--1960 & 15.44M & Public domain \\
2.& Wikipedia (QC category)        & Articles     & 2008--2025 & 31.15M & CC-BY-SA~3.0 \\
3.& CN2i -- \textit{Le Soleil}     & News articles& 2000--2025 & 5.86M  & Copyrighted \\ 
\hline
4.& CRIFUQ oral transcripts        & Interviews   & 2012--2019 & 0.76M  & Non-commercial \\
5.& Facebook \textit{Le Soleil}    & Comments     & 2020--2023 & 2.40M  & ToS-restricted \\
6.& Depotoir.ca                    & Forum posts  & 2009--2025 & 22.82M & Non-commercial \\
7.& MontrealRacing.com             & Forum posts  & 2003--2025 & 4.23M  & Non-commercial \\
8.& YouTube (QC channels)          & Comments     & 2006--2025 & 2.15M  & Non-commercial \\
9.& Reddit (QC communities)        & Comments     & 2025       & 1.76M  & Non-commercial \\ \midrule
& \multicolumn{3}{r}{\textbf{Total}} & \textbf{86.57M} & \\
\bottomrule
\end{tabular}}
\caption{Sources of the Qu\'{e}bec French corpus collected and used for CPT. Sizes are token counts using the CroissantLLM tokenizer.}
\label{tab:sources}
\end{table}

The data sources we used are reported in Table~\ref{tab:sources}. We can distinguish two categories: sources that include formal texts (numbers 1 to 3 in Table~\ref{tab:sources}) which make up 60\% (52.45M tokens) of our corpus, and those that include informal (spoken or user-generated) texts (numbers 4 to 9) which make up the remaining 40\% (34.12M tokens). These sources are described below: 

\begin{enumerate}

\item \textbf{BEQ Ebooks} (15.44M tokens):\\
Public-domain literary texts in Qu\'ebec French, sourced from the \textit{Bibliothèque électronique du Québec} (BEQ). These are primarily literary texts written in normative Qu\'ebec French. This source offers diachronic coverage of historical Qu\'ebecois literary language, enriching the overall corpus with stylistic and temporal diversity. 

\item \textbf{Wikipedia (QC)} (31.15M tokens):\\
Articles retrieved from the Qu\'{e}bec portal of Wikipedia, covering Qu\'{e}bec topics and presumed to be primarily written by Qu\'{e}b\'{e}cois people in formal prose. 
This source contributes broad, topic-specific coverage written in Qu\'ebec French.

\item \textbf{CN2i – \textit{Le Soleil} Newspaper} (5.86M tokens):\\
News articles from the Qu\'{e}bec City newspaper, \textit{Le Soleil}, written in formal journalistic prose. This source offers high-quality formal written texts covering a variety of current events.

\item \textbf{CRIFUQ Oral Transcripts} (0.76M tokens):\\
Interview transcriptions from the \textit{Centre de recherche interuniversitaire sur le français en usage au Québec} (CRIFUQ). It is our only collection of spontaneous speech in informal  Qu\'ebec French. 
It thus provides examples of features typical of Qu\'ebec French speech (e.g., ellipsis, disfluencies, phonology-to-orthography variation). 

\item \textbf{Facebook Comments – \textit{Le Soleil}} (2.40M tokens):\\
User comments on posts by the newspaper \textit{Le Soleil}, collected via the Facebook Graph API. These are examples of user-generated informal public discourse by newspaper-reading (thus older and more educated) individuals. It captures unedited conversational patterns and informal register. 

\item \textbf{Depotoir.ca} (22.82M tokens):\\
Public posts on the Qu\'ebec-based forum Depotoir.ca. These posts contain informal, colloquial, and regionally-specific Qu\'eb\'ecois, including slang and argot. This source contributes content in vernacular Qu\'ebec French rich in sociolinguistic variations, including non-standard orthography and expressive discourse, and covering a variety of cultural, social, and general discussion topics.

\item \textbf{MontrealRacing.com} (4.23M tokens):\\
Public posts from a Montreal-based automotive enthusiast forum. This source provides other examples of informal Qu\'ebec French, this time rich with technical terminology and slang.

\item \textbf{YouTube Comments} (2.15M tokens):\\
Public user comments from Qu\'ebec YouTube channels, collected via the YouTube API. These are texts of informal conversational Qu\'eb\'ecois, often containing regionalisms, and code-switching. They provide examples of more youthful writing patterns and slang.

\item \textbf{Reddit – Qu\'ebec Communities} (1.76M tokens):\\
Public comments from francophone Qu\'ebec subreddits, collected via Reddit API. Another example of informal, colloquial Qu\'ebec French, including slang and regional features. Unlike the other comments, this source provides examples of more sustained informal conversations. 
\end{enumerate}

Overall, our Qu\'{e}bec French corpus spans 86.57M tokens, which is significantly below typical CPT data budgets used for dialectal/low-resource adaptations, as shown in Table~\ref{tab:tokens}.

\subsection{Data Pre-Processing} 
\label{sec:data-preprocessing}

To use the Qu\'ebec French corpus for CPT, we applied light pre-processing to preserve dialectal markers and standardize formatting, allowing it to be used as a single, consistent dataset.
To that end, each line of text was stripped of byte-order markers, HTML or Wiki tags, and extra whitespaces. Empty lines were removed, but paragraph boundaries and sentence-level markers were preserved as much as possible. Only trivial corrections, such as collapsing repeated spaces, were applied, ensuring that distinct orthographic patterns were preserved.


\section{Training}
\label{sec:method}

 
To reduce the cost of CPT, we employed LoRA~\citep{hu2022lora} instead of updating all model parameters. Following common practice, we targeted the attention projections ($q$, $k$, $v$, $o$) and feed-forward layers ($\text{up}$, $\text{down}$, $\text{gate}$) with rank $r=16$, $\alpha=32$, and dropout of $0.1$. Trainable parameters are cast to \texttt{float32}, while frozen parameters remain in \texttt{float16}. Additionally, we enabled gradient checkpointing to further lower memory requirements. 
This strategy yielded a trainable parameter ratio of around $1\%$ of the full model, making dialectal adaptation feasible on modest hardware.

We performed CPT with a causal language modeling (CLM) objective. The model was exposed to our Qu\'{e}bec French corpus (see Section \ref{sec:data}) in a single pass without domain interleaving or replay of the previously used prestige French training data. Due to computational constraints, sequences were limited to $1024$ tokens with a stride of $512$, producing overlapping chunks that preserve context for long documents. We trained each model for 3 and 6 epochs over the corpus using the AdamW optimizer with a weight decay of $0.01$. The learning rate was set to $1\times 10^{-5}$ with cosine decay and a warm-up ratio of $0.1$. To avoid numerical instability during the training process, we clipped gradients at a norm of $1.0$.

The effective batch size in sequences can be described as:
\[
\text{EffBatch}_{\text{seq}} = b \times a \times d,
\]
where $b$ is the number of sequences per device (micro-batch), $a$ is the gradient-accumulation steps, and $d$ is the number of devices. 
With a sequence length of $1024$ tokens, the effective batch size in tokens was therefore $\text{EffBatch}_{\text{tok}} = \text{EffBatch}_{\text{seq}} \times 1024$. See Appendix~\ref{app:train-config} for more details. 

\section{Experimental Setup}
\label{sec:setup}

\subsection{Base Models}
We developed the Qu\'{e}bec French models by conducting CPT on the following base models:

\begin{itemize}
    \item \textbf{CroissantLLMChat-v0.1 (1.35B)}: a bilingual (English/French) open-source model. This serves as a strong dialect-aware baseline that is already exposed to French text.
    \item \textbf{Llama-3.2-1B}: a lightweight general-purpose model included as a baseline for multilingual models.
    \item \textbf{Llama-3.1-8B}: a high-capacity general-purpose model included as a stronger large-scale multilingual baseline for comparison.
\end{itemize}

All the models were evaluated both in their base forms and after 3 then 6 epochs of CPT following the training regime described in Section \ref{sec:method}. 

\subsection{Evaluation Benchmarks}
We evaluated our models on 8 tasks of the COLE French-language benchmark~\cite{beauchemin2025cole}.
To evaluate both the models' acquisition of Québec French and their retention of general abilities after CPT, we chose 4 Qu\'{e}bec French tasks (QFrCoLA, QFrBLiMP, QFrCoRE and QFrCoRT) and 4 prestige French tasks (AlloCin\'e, PAWS-X, Fr-BoolQ, and MMS). These 8 tasks are described below:

\begin{enumerate}
  \item \textbf{QFrCoLA}~\cite{beauchemin2025qfrcola}: 
  A  grammatical acceptability task. The dataset consists of a total of 25,153 individual sentences classified as grammatical or ungrammatical. The sentences are drawn from a normative Québec French resource, and grouped by phenomena (syntax, morphology, semantics, anglicism). We used the 7,546 test samples for evaluation.

  \item \textbf{QFrBLiMP}~\cite{beauchemin2025qfrblimp}: 
A  grammatical acceptability task. The dataset is composed of 1{,}761 carefully-edited sentence pairs, one correct and the other with a common linguistic mistake.  The
 task consists of identifying the correct sentence.
Linguistic mistakes cover 
 20 attested phenomena from the “Banque de dépannage linguistique” (BDL), an official Québec government grammar source, annotated by native speakers. We used the 529 test samples for evaluation.

\item  \textbf{QFrCoRE}~\cite{beauchemin2025set}: A  definition matching task. 
  The dataset includes 4,633  Qu\'ebec French multi-word expressions, with 10 possible definitions, and the correct one must be identified. The expressions and correct definitions were sourced from Québec regional-language collections. We used all 4,633 samples as a test set for evaluation.

  \item   \textbf{QFrCoRT}~\cite{beauchemin2025set}: A  definition matching task. 
  This task is similar to QFrCoRE, but for 201 single-word Qu\'ebec French idioms. We used all 201 samples as a test set for evaluation.


 

  \item \textbf{AlloCin\'e}~\cite{blard2020allocine}: 
  A binary sentiment classification of 200,000 movie reviews in French. We used the test set of 20,000 samples for evaluation. 

  \item \textbf{PAWS-X}~\cite{yang2019pawsx}: 
  A paraphrase detection task. This task consists in identifying sentences that are paraphrases of each other. This multilingual dataset is composed of 23,659 professionally translated sentence pairs, but we used the test set of 2,000 French samples for evaluation. 

  \item \textbf{Fr-BoolQ}~\cite{clark2019boolq}: 
  A binary reading comprehension task. This task is composed of 15,942 yes/no questions paired with passages containing the answer. This dataset is also multilingual, but we used the French subset of 178 test samples. 

  \item \textbf{MMS}~\cite{brand24mmsbenchmark}: 
  A sentiment analysis task. This task provides text snippets taken from 79 corpora from various domains, and requires identifying the sentiment in a binary- or trinary-class schema, depending on the corpus. This dataset is also multilingual, but we used the French subset of 63,190 test samples.
\end{enumerate}

Table~\ref{tab:label-freq} summarizes all of these tasks with respect to their label distribution.

\begin{table}[!t]
\centering
\scriptsize
\setlength{\tabcolsep}{3pt}
\begin{tabular}{@{}crrrrrrrr@{}}
\toprule
& \multicolumn{4}{c}{\textbf{Qu\'eb\'ecois}} & \multicolumn{4}{c}{\textbf{French}} \\
\cmidrule(lr){2-5}\cmidrule(lr){6-9}
\textbf{Label} &
\rotatebox{90}{QFrCoLA} &
\rotatebox{90}{QFrBLiMP} &
\rotatebox{90}{QFrCoRE} &
\rotatebox{90}{QFrCoRT} &
\rotatebox{90}{Allocin\'e} &
\rotatebox{90}{Fr-BoolQ} &
\rotatebox{90}{MMS} &
\rotatebox{90}{PAWS-X} \\
\midrule
0 & 30.5 & 48.0 & 10.7 & 12.8 & 52.0 & 50.0 & 39.9 & 54.8 \\
1 & 69.5 & 52.0 &  9.9 &  8.1 & 47.9 & 50.0 & 20.5 & 45.1 \\
2 & --   & --   &  9.0 & 11.1 & --   & --   & 39.5 & --   \\
3 & --   & --   & 10.2 &  9.3 & --   & --   & --   & --   \\
4 & --   & --   & 10.1 &  7.6 & --   & --   & --   & --   \\
5 & --   & --   & 10.3 & 10.5 & --   & --   & --   & --   \\
6 & --   & --   &  9.3 & 11.6 & --   & --   & --   & --   \\
7 & --   & --   &  9.6 & 10.5 & --   & --   & --   & --   \\
8 & --   & --   & 10.3 &  9.3 & --   & --   & --   & --   \\
9 & --   & --   & 10.1 &  8.7 & --   & --   & --   & --   \\
\bottomrule
\end{tabular}
\caption{Label frequency (\%) across the selected COLE tasks.}
\label{tab:label-freq}
\end{table}

\begin{table*}[!t]
\centering
\small
\setlength{\tabcolsep}{5pt}
\renewcommand{\arraystretch}{1.1}
\resizebox{\textwidth}{!}{%
\begin{tabular}{ll|cc|cc|cc|cc|c|}
\cline{3-11}
 & &
\multicolumn{9}{c|}{\textbf{COLE Québec-French Tasks}} \\
\hline
\multicolumn{1}{|l|}{\textbf{Methods}} & \textbf{Model} &
\multicolumn{2}{c}{\textbf{QFrCoLA}} &
\multicolumn{2}{c}{\textbf{QFrBLiMP}} &
\multicolumn{2}{c}{\textbf{QFrCoRE}} &
\multicolumn{2}{c|}{\textbf{QFrCoRT}} &
\cellcolor{grey}{\(\Delta\)\textbf{QC-FR}} \\
\multicolumn{1}{|l|}{} & &
\textbf{macroF1} & \(\Delta\)F1 &
\textbf{macroF1} & \(\Delta\)F1 &
\textbf{macroF1} & \(\Delta\)F1 &
\textbf{macroF1} & \(\Delta\)F1 &
\cellcolor{grey}{\textbf{avgF1}} \\
\hline
\multicolumn{2}{|l|}{\textbf{CroissantLLM family}} & \multicolumn{9}{c|}{} \\
\multicolumn{1}{|l}{} & Croissant (Base)
& $27.33$ & -- & $32.84$ & -- & $2.91$ & -- & $4.59$ & -- & \cellcolor{grey}{--} \\
\multicolumn{1}{|l}{} & Croissant (3 epochs CPT)
& $26.56$ & $-0.77$ & $32.84$ & $0.00$ & $2.77$ & $-0.14$ & $5.87$ & $+1.28$ & \cellcolor{grey}{$+0.09$} \\
\multicolumn{1}{|l}{} & Croissant (6 epochs CPT)
& $\boxed{46.22}$ & $+18.89$ & $\boxed{35.90}$ & $+3.06$ & $1.76$ & $-1.15$ & $1.61$ & $-2.98$ & \cellcolor{grey}{$+4.45$} \\
\hline
\multicolumn{2}{|l|}{\textbf{Llama-3.2-1B family}} & \multicolumn{9}{c|}{} \\
\multicolumn{1}{|l}{} & Llama-1B (Base)
& $45.84$ & -- & $32.44$ & -- & $1.83$ & -- & $2.00$ & -- & \cellcolor{grey}{--} \\
\multicolumn{1}{|l}{} & Llama-1B (3 epochs CPT)
& $27.89$ & $-17.95$ & $34.92$ & $+2.45$ & $3.68$ & $+1.85$ & $4.54$ & $+2.54$ & \cellcolor{grey}{$-2.78$} \\
\multicolumn{1}{|l}{} & Llama-1B (6 epochs CPT)
& $42.45$ & $-3.39$ & $\boxed{35.90}$ & $+3.46$ & $4.58$ & $+2.75$ & $3.88$ & $+1.88$ & \cellcolor{grey}{$+1.18$} \\
\hline
\multicolumn{2}{|l|}{\textbf{Llama-3.1-8B family}} & \multicolumn{9}{c|}{} \\
\multicolumn{1}{|l}{} & Llama-8B (Base)
& $41.04$ & -- & $32.44$ & -- & $5.84$ & -- & $10.54$ & -- & \cellcolor{grey}{--} \\
\multicolumn{1}{|l}{} & Llama-8B (3 epochs CPT)
& $40.99$ & $-0.05$ & $32.44$ & $0.00$ & $8.12$ & $+2.28$ & $12.33$ & $+1.79$ & \cellcolor{grey}{$+1.01$} \\
\multicolumn{1}{|l}{} & Llama-8B (6 epochs CPT)
& $41.66$ & $+0.62$ & $32.44$ & $0.00$ & $\boxed{8.91}$ & $+3.07$ & $\boxed{13.73}$ & $+3.19$ & \cellcolor{grey}{$+1.72$} \\
\hline
\end{tabular}}
\caption{Results on COLE Qu\'{e}bec French tasks only. \(\boxed{Boxed}\) marks global best within each column. \(\Delta\) columns show improvement over the corresponding base model; gray cells summarize the average \(\Delta\) across Qu\'{e}bec French tasks.}
\label{tab:cole-qcfr}
\end{table*}

\begin{table*}[!t]
\centering
\small
\setlength{\tabcolsep}{5pt}
\renewcommand{\arraystretch}{1.1}
\resizebox{\textwidth}{!}{%
\begin{tabular}{ll|cc|cc|cc|cc|c|}
\cline{3-11}
 & &
\multicolumn{9}{c|}{\textbf{General French Tasks}} \\
\hline
\multicolumn{1}{|l|}{\textbf{Methods}} & \textbf{Model} &
\multicolumn{2}{c}{\textbf{AlloCiné}} &
\multicolumn{2}{c}{\textbf{PAWS-X}} &
\multicolumn{2}{c}{\textbf{Fr-BoolQ}} &
\multicolumn{2}{c|}{\textbf{MMS}} &
\cellcolor{grey}{\(\Delta\)\textbf{FR}} \\
\multicolumn{1}{|l|}{} & &
\textbf{macroF1} & \(\Delta\)F1 &
\textbf{macroF1} & \(\Delta\)F1 &
\textbf{macroF1} & \(\Delta\)F1 &
\textbf{macroF1} & \(\Delta\)F1 &
\cellcolor{grey}{\textbf{avgF1}} \\
\hline
\multicolumn{2}{|l|}{\textbf{CroissantLLM family}} & \multicolumn{9}{c|}{} \\
\multicolumn{1}{|l}{} & Croissant (Base)
& $36.18$ & -- & $40.57$ & -- & $\boxed{50.75}$ & -- & $19.37$ & -- & \cellcolor{grey}{--} \\
\multicolumn{1}{|l}{} & Croissant (3 epochs CPT)
& $41.10$ & $+4.92$ & $41.55$ & $+0.98$ & $44.30$ & $-6.45$ & $26.42$ & $+7.05$ & \cellcolor{grey}{$+1.63$} \\
\multicolumn{1}{|l}{} & Croissant (6 epochs CPT)
& $50.39$ & $+14.21$ & $\boxed{50.07}$ & $+9.50$ & $32.06$ & $-18.69$ & $28.51$ & $+9.14$ & \cellcolor{grey}{$+3.54$} \\
\hline
\multicolumn{2}{|l|}{\textbf{Llama-3.2-1B family}} & \multicolumn{9}{c|}{} \\
\multicolumn{1}{|l}{} & Llama-1B (Base)
& $\boxed{70.96}$ & -- & $35.42$ & -- & $50.29$ & -- & $19.11$ & -- & \cellcolor{grey}{--} \\
\multicolumn{1}{|l}{} & Llama-1B (3 epochs CPT)
& $45.05$ & $-25.91$ & $45.11$ & $+9.69$ & $40.80$ & $-9.49$ & $19.89$ & $+0.78$ & \cellcolor{grey}{$-6.23$} \\
\multicolumn{1}{|l}{} & Llama-1B (6 epochs CPT)
& $49.77$ & $-21.19$ & $35.56$ & $+0.14$ & 3$9.21$ & $-11.08$ & $19.12$ & $+0.01$ & \cellcolor{grey}{$-8.03$} \\
\hline
\multicolumn{2}{|l|}{\textbf{Llama-3.1-8B family}} & \multicolumn{9}{c|}{} \\
\multicolumn{1}{|l}{} & Llama-8B (Base)
& $34.37$ & -- & $34.10$ & -- & $40.20$ & -- & $22.17$ & -- & \cellcolor{grey}{--} \\
\multicolumn{1}{|l}{} & Llama-8B (3 epochs CPT)
& $53.95$ & $+19.58$ & $47.89$ & $+13.79$ & $39.13$ & $-1.07$ & $30.82$ & $+8.65$ & \cellcolor{grey}{$+10.24$} \\
\multicolumn{1}{|l}{} & Llama-8B (6 epochs CPT)
& $57.11$ & $+22.74$ & $49.81$ & $+15.71$ & $38.17$ & $-2.03$ & \boxed{37.82} & $+15.65$ & \cellcolor{grey}{$+13.02$} \\
\hline
\end{tabular}}
\caption{Results on general French tasks only. \(\boxed{Boxed}\) marks global best within each column. \(\Delta\) columns show improvement over the corresponding base model; gray cells summarize the average \(\Delta\) across prestige French tasks.}
\label{tab:cole-fr}
\end{table*}

None of the texts in the Québec French tasks (QFrCoLA, QFrBLiMP, QFrCoRE, QFrCoRT) were built from, or were included in the Québec French unlabeled corpus used for CPT (see Section~\ref{sec:data-sources}).




\section{Results and Analysis}
\label{sec:analysis}
Tables~\ref{tab:cole-qcfr} and~\ref{tab:cole-fr} shows the macro-F1 for each model and for each task, along with the variation in macro-F1 ($\Delta$F1) between the base model and the CPT version. In addition, the tables provide the CPT model's average change in macro-F1 over all four Qu\'{e}bec French tasks and all four general French tasks (grey columns). We chose to use macro-F1 to account for the label imbalance that exists in some datasets (see Table~\ref{tab:label-freq}). 

We analyze the results on three aspects. First, we consider the models' ability to acquire Qu\'{e}b\'{e}cois (see Section~\ref{sec:acquisition}). Second, we look at the models' retention of prestige French abilities (see Section~\ref{sec:gen-forgetting}). Finally, we study the adaptation-retention trade-off of the models, or their ability to balance learning and remembering language (see Section~\ref{sec:tradeoff}).

\subsection{Qu\'{e}bec French Acquisition}
\label{sec:acquisition}

\begin{figure}
    \centering
    \includegraphics[width=1.0\linewidth]{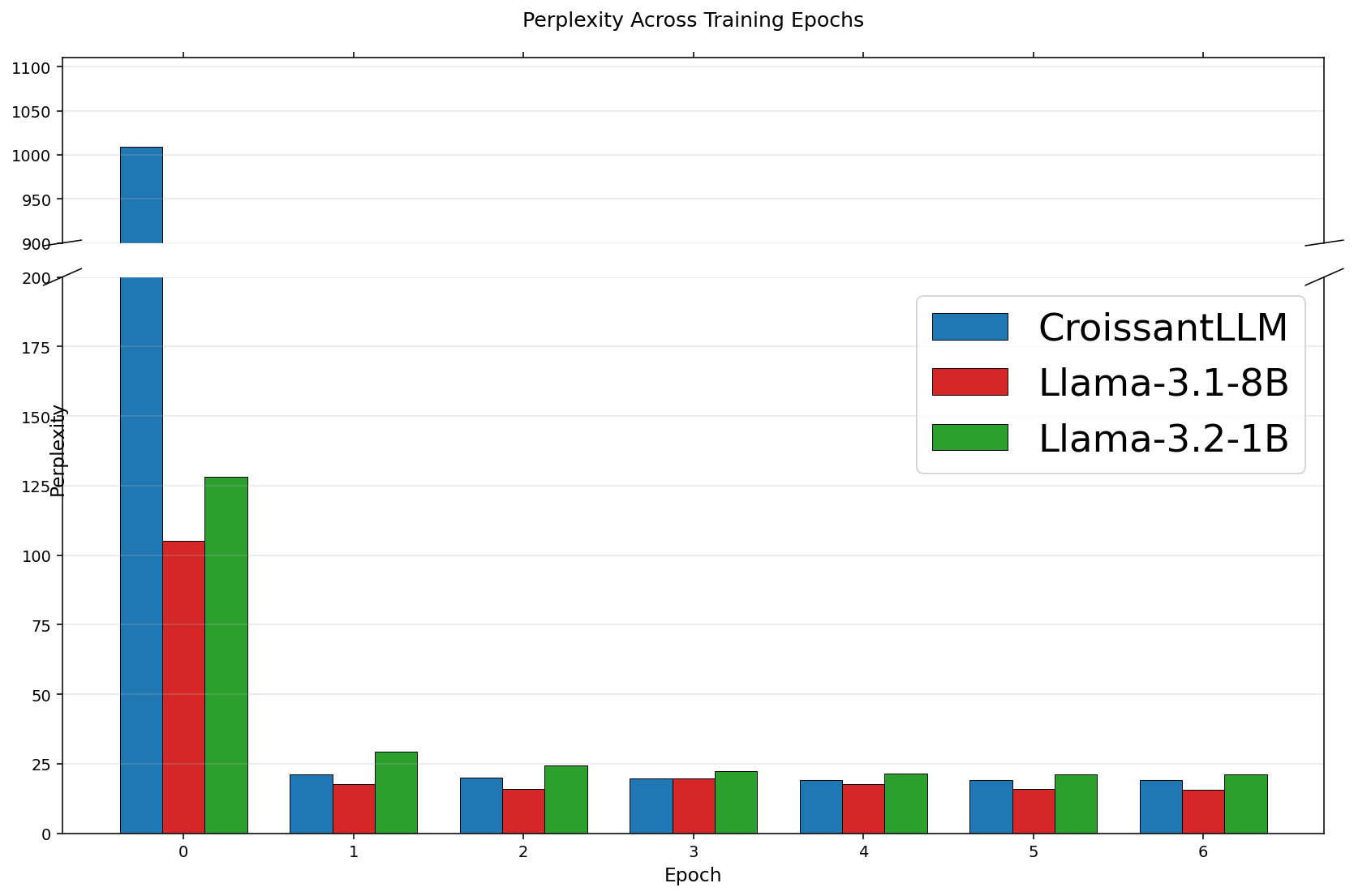}
    \caption{Perplexity during CPT training. Lower perplexity indicates better fit to Québec French. All models are evaluated on identical held-out corpus.}
    \label{fig:perplexity}
\end{figure}

Figure~\ref{fig:perplexity} shows the perplexity of the three models before (epoch 0) and after each of the six epochs of CPT. To compute perplexity scores, we used 5\% of the COLE test sets during training. Because this is training perplexity (with overlap), it likely underestimates true generalization error. We therefore treat this perplexity as a proxy for adaptation dynamics rather than a calibrated held-out metric, and leave non-overlapping held-out perplexity to future work. Nonetheless, the results show that all three models start with high training perplexity, consistent with a distribution gap between their pre-training mix and Qu\'{e}bec French, and drop sharply after the first epoch, indicating rapid uptake of dialectal patterns under CPT.

The results in Table~\ref{tab:cole-qcfr} show that all three models improve their performances in Qu\'{e}bec French tasks after 6 epochs of CPT. However, these improvements are not uniform across tasks nor models. Out of all four tasks, QFrCoLA proves to be the most difficult one, and most models actually worsen on it or show very small gains after CPT. This may be because of a conflict between the task's objectives and the training data we used. Indeed, the task consists in labeling a sentence as grammatical or not according to normative Qu\'{e}bec French rules; the ungrammatical sentences contain linguistic mistakes that are common in Qu\'{e}bec. On the other hand, much of our Qu\'{e}bec training data (40\%) comes from unedited sources, such as web forum posts and online comments, where these mistakes would be frequently found. The models have thus been trained to accept these erroneous phrasings as ``correct''. This would warrant future investigation by conducting CPT exclusively on more formal Qu\'ebec French sources.

We can note as well that there is a second normative Qu\'{e}bec French task, QFrBLiMP, on which our models actually improve after CPT. The difference is that this task is a binary comparison, where the model is given two sentences and must identify the correct from the incorrect one. Taken together, these results indicate that CPT is making the models better at understanding both normative and slang Qu\'{e}bec French: the models perform better at distinguishing normative from incorrect language, but also more accepting of incorrect popular language.

Training time seems correlated with performance, as all three models show more improvement after 6 epochs of CPT than 3. However, the correlation to model size is less clear. Llama-3.1-8B clearly outperforms Llama-3.2-1B, but is in turn outperformed by CroissantLLM with 1.35B parameters. 

\subsection{Prestige French Retention}
\label{sec:gen-forgetting}

As shown in Table~\ref{tab:cole-fr}, two out of the three  models, CroissantLLM and Llama-3.1-8B, not only retained their performances on the prestige French tasks but actually improved their average performance, showing that training on a regional dialect helps in handling the prestige dialect as well. Only the smallest model, Llama-3.2-1B, seems to be forgetting general French skills with CPT, and actually becomes worse with more epochs.

The Fr-BoolQ task is the only one whose performance suffers from CPT with all three models. This is probably because our training dataset has no question-answering sources. This, combined with the fact CPT adapts models without replay or regularization, seems to have progressively overwritten the specialized skills needed for this task. However, the other tasks, paraphrase recognition and sentiment detection, require more general language-understanding skills and thus benefit from the expanded language understanding that CPT with Qu\'{e}bec French provides (at least for the two models that benefit from CPT).


\subsection{Adaptation-Retention Trade-off}
\label{sec:tradeoff}
The results in Tables~\ref{tab:cole-qcfr} and \ref{tab:cole-fr} show that the smaller models, CroissantLLM and Llama-3.2-1B, have difficulty balancing adapting to a new dialect and retaining the skills they have learned previously. CroissantLLM does improve in both Qu\'{e}bec French understanding and prestige French ability, but the gains are modest after 3 epochs and the greater gains in the former after 6 epochs almost completely wipe out the gains in the latter. Meanwhile, Llama-3.2-1B's performance degrades overall throughout the experiment. On the other hand, Llama-3.1-8B, our largest model, shows clear improvements in both Qu\'{e}bec French understanding and prestige French ability, and these benefits increase with the number of training epochs. This demonstrates that there is value to using CPT to train models for regional dialects, but only if the base model is large enough to absorb the new information without losing the knowledge gained from its initial training.

\section{Conclusion}
\label{sec:conclusion}

In this paper, we demonstrated a case study of low-resource dialectal adaptation of LLMs on Qu\'{e}bec French.
Our results on 8 tasks from the French benchmark suite COLE show that CPT with LoRA achieves substantial dialectal gains and also improves the model performance on prestige French tasks, but only if the model is large enough to support it. 
Our work also demonstrated that this specialization can be made by updating $\leq\!1\%$ of model parameters and with using a very small corpus (85M tokens), but that corpus composition has an important impact. Indeed, our corpus being rich in informal and unedited sources, and lacking question-answering sources made the trained models less proficient at distinguishing normatively correct and incorrect text and at question-answering. 



As future work, we could explore varying the mix of data sources and examine their influence on performance across specific benchmarks. Further research could also investigate cross-dialect and cross-language CPT, including scenarios that involve code-switching or mixed registers. Finally, exploring techniques such as selective parameter freezing may offer a way to enhance language retention.

\section{Societal Impact and Ethical Considerations}
\label{sec:ethics}

\paragraph{Linguistic equity and preservation.}
Unlike general LLMs which are focused on more popular high-resource languages, our work is designed to tailor LLMs to low-resource regional dialects such as Qu\'{e}bec French, providing more equitable access to AI tools.

\paragraph{Representation biases and stereotype risks.}
Training on naturally occurring Qu\'{e}bec slang can strenghten stereotypes linking dialectal features to class, region, or demographics. Our corpus spans news, social media, and forums to diversify exposure, but skews toward written, urban, younger, internet-active speakers while under-representing rural communities, older generations, Indigenous speakers, immigrants, and spontaneous speech. Model behavior may not generalize across these groups. Claims about ``Québec French'' should be read as conditional on training data demographics.

\paragraph{Dialect subordination via ``correction.''}
Systems defaulting to prestige French as ``correct'' (grammar checkers, translation, autocomplete) can normalize away regional features and reinforce linguistic hierarchies. We advocate designs that preserve dialectal variation and distinguish appropriateness from error.

\section*{Acknowledgments}
This project was undertaken thanks to funding from NSERC, IVADO and the Canada First Research Excellence Fund. 
The authors would also like to thank CN2i / Le Soleil for generously giving us access to their data. 

\section{Limitations}
\label{sec:limitations}

\paragraph{Benchmark coverage and task selection.}
Our evaluation focuses exclusively on a subset of the COLE benchmark's language tasks, which emphasize grammatical acceptability and linguistic structure. This scope excludes sociolinguistic variation (register, formality), conversational competence (authentic dialogue, code-switching), generation quality (dialectal fluency, idiom production), and domain-specific tasks (legal, medical terminology). Human evaluation of generated text would provide critical insights into dialectal authenticity that our automated metrics cannot capture.

\paragraph{Ablation studies and design choices.}
We do not report systematic ablations on key hyperparameters: LoRA rank sensitivity ($r{=}16$ was used following common practice without exploring further values), layer targeting (we apply LoRA to all attention and FFN layers without testing attention-only or selective configurations), gradient checkpointing impact, or adaptive versus uniform sampling strategies. These ablations would strengthen support for our design decisions but were omitted due to computational constraints.

\paragraph{Computational constraints.}
Our experiments use modest hardware (single/dual V100 GPUs) with 1024-token sequences, potentially constraining discourse-level phenomena, hyperparameter exploration, and training beyond 6 epochs. While our resource-efficient focus enables broader accessibility, it limits the performance ceiling we can explore.

\paragraph{Access to  Qu\'{e}bec French data.}
High-quality  Qu\'{e}bec French text is hard to find, scattered, and often behind legal or platform restrictions. Licenses and Terms of Service limit what we can share (e.g., news APIs, social-media comments), and speech transcripts are small. Informal writing exists but is uneven across topics, regions, and age groups. These limits keep our CPT small, make full reproducibility difficult (some sources cannot be released), and may bias models toward urban, online, younger speakers. 

\section{Bibliographical References}
\bibliographystyle{lrec}
\bibliography{bibliography}

\begin{thebibliography}{44}
\providecommand{\natexlab}[1]{#1}

\bibitem[{Aljundi et~al.(2018)Aljundi, Babiloni, Elhoseiny, Rohrbach, and Tuytelaars}]{aljundi2018memoryawaresynapseslearning}
Rahaf Aljundi, Francesca Babiloni, Mohamed Elhoseiny, Marcus Rohrbach, and Tinne Tuytelaars. 2018.
\newblock Memory aware synapses: Learning what (not) to forget.
\newblock In \emph{Proceedings of the European Conference on Computer Vision (ECCV)}, pages 139--154, Munich, Germany. Springer.

\bibitem[{Augustyniak et~al.(2023)Augustyniak, Wo{\'z}niak, Gruza, Gramacki, Rajda, Morzy, and Kajdanowicz}]{brand24mmsbenchmark}
{\L}ukasz Augustyniak, Szymon Wo{\'z}niak, Marcin Gruza, Piotr Gramacki, Krzysztof Rajda, Miko{\l}aj Morzy, and Tomasz~Jan Kajdanowicz. 2023.
\newblock \href {https://openreview.net/forum?id=7tMgzSvopH} {Massively multilingual corpus of sentiment datasets and multi-faceted sentiment classification benchmark}.
\newblock In \emph{Proceedings of the Thirty-seventh Conference on Neural Information Processing Systems (NeurIPS 2023) Datasets and Benchmarks Track}.

\bibitem[{Bari et~al.(2025)Bari, Alnumay, Alzahrani, Alotaibi, Alyahya, AlRashed, Mirza, Alsubaie, Alahmed, Alabduljabbar, Alkhathran, Almushayqih, Alnajim, Alsubaihi, Al~Mansour, Hassan, Alrubaian, Alammari, Alawami, Al-Thubaity, Abdelali, Kuriakose, Abujabal, Al-Twairesh, Alowisheq, and Khan}]{allam2024}
M.~Saiful Bari, Yazeed Alnumay, Norah~A. Alzahrani, Nouf~M. Alotaibi, Hisham~Abdullah Alyahya, Sultan AlRashed, Faisal~Abdulrahman Mirza, Shaykhah~Z. Alsubaie, Hassan~A. Alahmed, Ghadah Alabduljabbar, Raghad Alkhathran, Yousef Almushayqih, Raneem Alnajim, Salman Alsubaihi, Maryam Al~Mansour, Saad~Amin Hassan, Majed Alrubaian, Ali Alammari, Zaki Alawami, and 7 others. 2025.
\newblock \href {https://openreview.net/forum?id=MscdsFVZrN} {Allam: Large language models for arabic and english}.
\newblock In \emph{Proceedings of the Thirteenth International Conference on Learning Representations (ICLR)}.

\bibitem[{Beauchemin and Khoury(2025)}]{beauchemin2025qfrcola}
David Beauchemin and Richard Khoury. 2025.
\newblock \href {https://doi.org/10.18653/v1/2025.emnlp-main.6} {{QF}r{C}o{LA}: a {Q}uebec-{F}rench corpus of linguistic acceptability judgments}.
\newblock In \emph{Proceedings of the 2025 Conference on Empirical Methods in Natural Language Processing}, pages 119--130, Suzhou, China. Association for Computational Linguistics.

\bibitem[{Beauchemin et~al.(2025{\natexlab{a}})Beauchemin, Tremblay, Youssef, and Khoury}]{beauchemin2025cole}
David Beauchemin, Yan Tremblay, Mohamed~Amine Youssef, and Richard Khoury. 2025{\natexlab{a}}.
\newblock \href {https://arxiv.org/abs/2510.05046} {Cole: A comprehensive benchmark for french language understanding evaluation}.
\newblock \emph{Preprint}, arXiv:2510.05046.

\bibitem[{Beauchemin et~al.(2025{\natexlab{b}})Beauchemin, Tremblay, Youssef, and Khoury}]{beauchemin2025set}
David Beauchemin, Yan Tremblay, Mohamed~Amine Youssef, and Richard Khoury. 2025{\natexlab{b}}.
\newblock \href {https://arxiv.org/abs/2510.05026} {A set of quebec-french corpora of regional expressions and terms}.
\newblock \emph{Preprint}, arXiv:2510.05026.

\bibitem[{Beauchemin et~al.(2025{\natexlab{c}})Beauchemin, Veilleux, Khoury, and Roy}]{beauchemin2025qfrblimp}
David Beauchemin, Pier-Luc Veilleux, Richard Khoury, and Johanna-Pascale Roy. 2025{\natexlab{c}}.
\newblock \href {https://arxiv.org/abs/2509.25664} {Qfrblimp: A quebec-french benchmark of linguistic minimal pairs}.
\newblock \emph{Preprint}, arXiv:2509.25664.

\bibitem[{Beltagy et~al.(2019)Beltagy, Lo, and Cohan}]{beltagy2019scibert}
Iz~Beltagy, Kyle Lo, and Arman Cohan. 2019.
\newblock \href {https://doi.org/10.18653/v1/D19-1371} {{S}ci{BERT}: A pretrained language model for scientific text}.
\newblock In \emph{Proceedings of the 2019 Conference on Empirical Methods in Natural Language Processing and the 9th International Joint Conference on Natural Language Processing (EMNLP-IJCNLP)}, pages 3615--3620, Hong Kong, China. Association for Computational Linguistics.

\bibitem[{Ben~Zaken et~al.(2022)Ben~Zaken, Goldberg, and Ravfogel}]{ben-zaken-etal-2022-bitfit}
Elad Ben~Zaken, Yoav Goldberg, and Shauli Ravfogel. 2022.
\newblock \href {https://doi.org/10.18653/v1/2022.acl-short.1} {{B}it{F}it: Simple parameter-efficient fine-tuning for transformer-based masked language-models}.
\newblock In \emph{Proceedings of the 60th Annual Meeting of the Association for Computational Linguistics (Volume 2: Short Papers)}, pages 1--9, Dublin, Ireland. Association for Computational Linguistics.

\bibitem[{Blard(2020)}]{blard2020allocine}
Th{\'e}ophile Blard. 2020.
\newblock French sentiment analysis with bert (allocin{\'e} dataset).
\newblock \url{https://github.com/TheophileBlard/french-sentiment-analysis-with-bert}.
\newblock Allocin{\'e}: 200k French movie reviews for sentiment analysis.

\bibitem[{{CataLlama Team}(2024)}]{cataLlama2024}
{CataLlama Team}. 2024.
\newblock Catallama: Llama 3 models for catalan.
\newblock \url{https://huggingface.co/catallama}.
\newblock Hugging Face model repository.

\bibitem[{Chalkidis et~al.(2020)Chalkidis, Fergadiotis, Malakasiotis, Aletras, and Androutsopoulos}]{chalkidis-etal-2020-legal}
Ilias Chalkidis, Manos Fergadiotis, Prodromos Malakasiotis, Nikolaos Aletras, and Ion Androutsopoulos. 2020.
\newblock \href {https://doi.org/10.18653/v1/2020.findings-emnlp.261} {{LEGAL-BERT}: The muppets straight out of law school}.
\newblock In \emph{Findings of the Association for Computational Linguistics: EMNLP 2020}, pages 2898--2904, Online. Association for Computational Linguistics.

\bibitem[{Chaudhry et~al.(2019)Chaudhry, Rohrbach, Elhoseiny, Ajanthan, Dokania, Torr, and Ranzato}]{chaudhry2019tinyepisodicmemoriescontinual}
Arslan Chaudhry, Marcus Rohrbach, Mohamed Elhoseiny, Thalaiyasingam Ajanthan, Puneet~K. Dokania, Philip H.~S. Torr, and Marc{'}Aurelio Ranzato. 2019.
\newblock \href {https://arxiv.org/abs/1902.10486} {On tiny episodic memories in continual learning}.
\newblock \emph{Preprint}, arXiv:1902.10486.

\bibitem[{Chronopoulou et~al.(2019)Chronopoulou, Baziotis, and Potamianos}]{chronopoulou-etal-2019-embarrassingly}
Alexandra Chronopoulou, Christos Baziotis, and Alexandros Potamianos. 2019.
\newblock \href {https://doi.org/10.18653/v1/N19-1213} {An embarrassingly simple approach for transfer learning from pretrained language models}.
\newblock In \emph{Proceedings of the 2019 Conference of the North {A}merican Chapter of the Association for Computational Linguistics: Human Language Technologies, Volume 1 (Long and Short Papers)}, pages 2089--2095, Minneapolis, Minnesota. Association for Computational Linguistics.

\bibitem[{Clark et~al.(2019)Clark, Lee, Chang, Kwiatkowski, Collins, and Toutanova}]{clark2019boolq}
Christopher Clark, Kenton Lee, Ming-Wei Chang, Tom Kwiatkowski, Michael Collins, and Kristina Toutanova. 2019.
\newblock \href {https://aclanthology.org/N19-1300/} {Boolq: Exploring the surprising difficulty of natural yes/no questions}.
\newblock In \emph{Proceedings of the 2019 Conference of the North American Chapter of the Association for Computational Linguistics: Human Language Technologies (NAACL-HLT 2019)}.

\bibitem[{Conneau et~al.(2020)Conneau, Khandelwal, Goyal, Chaudhary, Wenzek, Guzm{\'a}n, Grave, Ott, Zettlemoyer, and Stoyanov}]{conneau-etal-2020-unsupervised}
Alexis Conneau, Kartikay Khandelwal, Naman Goyal, Vishrav Chaudhary, Guillaume Wenzek, Francisco Guzm{\'a}n, Edouard Grave, Myle Ott, Luke Zettlemoyer, and Veselin Stoyanov. 2020.
\newblock \href {https://doi.org/10.18653/v1/2020.acl-main.747} {Unsupervised cross-lingual representation learning at scale}.
\newblock In \emph{Proceedings of the 58th Annual Meeting of the Association for Computational Linguistics (ACL 2020)}, pages 8440--8451, Online. Association for Computational Linguistics.

\bibitem[{Dabre et~al.(2025)Dabre, Puduppully, Suzuki et~al.}]{dabre2025nemotron}
Raj Dabre, Ratish Puduppully, Akihiro Suzuki, and 1 others. 2025.
\newblock \href {https://aclanthology.org/2025.indonlp-1.6/} {Adapting multilingual llms to low-resource languages using continued pre-training and synthetic corpora: A case study for hindi llms}.
\newblock In \emph{Proceedings of the 6th Workshop on Indonesian Language Processing (IndoNLP 2025)}, pages 53--67.

\bibitem[{Dettmers et~al.(2023)Dettmers, Pagnoni, Holtzman, and Zettlemoyer}]{dettmers2023qlora}
Tim Dettmers, Artidoro Pagnoni, Ari Holtzman, and Luke Zettlemoyer. 2023.
\newblock \href {https://proceedings.neurips.cc/paper_files/paper/2023/file/1feb87871436031bdc0f2beaa62a049b-Paper-Conference.pdf} {{QL}o{RA}: Efficient finetuning of quantized {LLM}s}.
\newblock In \emph{Advances in Neural Information Processing Systems}, volume~36, pages 10088--10115.

\bibitem[{Elhady et~al.(2025)Elhady, Agirre, and Artetxe}]{elhady2025emergent}
Ahmed Elhady, Eneko Agirre, and Mikel Artetxe. 2025.
\newblock \href {https://doi.org/10.18653/v1/2025.acl-long.1547} {Emergent abilities of large language models under continued pre-training for language adaptation}.
\newblock In \emph{Proceedings of the 63rd Annual Meeting of the Association for Computational Linguistics (Volume 1: Long Papers)}, pages 32174--32186, Vienna, Austria. Association for Computational Linguistics.

\bibitem[{Etxaniz et~al.(2024)Etxaniz, Sainz, Perez, Aldabe, Rigau, Agirre, Ormazabal, Artetxe, and Soroa}]{agerri2024latxa}
Julen Etxaniz, Oscar Sainz, Naiara Perez, Itziar Aldabe, German Rigau, Eneko Agirre, Aitor Ormazabal, Mikel Artetxe, and Aitor Soroa. 2024.
\newblock \href {https://doi.org/10.18653/v1/2024.acl-long.799} {Latxa: An open language model and evaluation suite for basque}.
\newblock In \emph{Proceedings of the 62nd Annual Meeting of the Association for Computational Linguistics (Volume 1: Long Papers)}, pages 14952--14972, Bangkok, Thailand. Association for Computational Linguistics.

\bibitem[{Farajtabar et~al.(2020)Farajtabar, Azizan, Mott, and Li}]{farajtabar2020orthogonal}
Mehrdad Farajtabar, Navid Azizan, Alex Mott, and Ang Li. 2020.
\newblock \href {https://proceedings.mlr.press/v108/farajtabar20a.html} {Orthogonal gradient descent for continual learning}.
\newblock In \emph{Proceedings of the 23rd International Conference on Artificial Intelligence and Statistics}, volume 108 of \emph{Proceedings of Machine Learning Research}, pages 3762--3773, Online. PMLR.

\bibitem[{Fishel et~al.(2025)Fishel, Tuisk, and Alum{\"a}e}]{turtunlp2025sorbian}
Mark Fishel, Taido Tuisk, and Tanel Alum{\"a}e. 2025.
\newblock \href {https://www2.statmt.org/wmt25/pdf/2025.wmt-1.88.pdf} {Tartunlp at wmt25: Llms with limited resources for slavic languages}.
\newblock In \emph{Proceedings of the Tenth Conference on Machine Translation (WMT 2025)}.

\bibitem[{Gururangan et~al.(2020)Gururangan, Marasovi{\'c}, Swayamdipta, Lo, Beltagy, Downey, and Smith}]{gururangan-etal-2020-dont}
Suchin Gururangan, Ana Marasovi{\'c}, Swabha Swayamdipta, Kyle Lo, Iz~Beltagy, Doug Downey, and Noah~A. Smith. 2020.
\newblock \href {https://doi.org/10.18653/v1/2020.acl-main.740} {Don{'}t stop pretraining: Adapt language models to domains and tasks}.
\newblock In \emph{Proceedings of the 58th Annual Meeting of the Association for Computational Linguistics}, pages 8342--8360, Online. Association for Computational Linguistics.

\bibitem[{Houlsby et~al.(2019)Houlsby, Giurgiu, Jastrzebski, Morrone, De~Laroussilhe, Gesmundo, Attariyan, and Gelly}]{houlsby2019parameter}
Neil Houlsby, Andrei Giurgiu, Stanislaw Jastrzebski, Bruna Morrone, Quentin De~Laroussilhe, Andrea Gesmundo, Mona Attariyan, and Sylvain Gelly. 2019.
\newblock \href {https://proceedings.mlr.press/v97/houlsby19a.html} {Parameter-efficient transfer learning for {NLP}}.
\newblock In \emph{Proceedings of the 36th International Conference on Machine Learning}, volume~97 of \emph{Proceedings of Machine Learning Research}, pages 2790--2799, Long Beach, California, USA. PMLR.

\bibitem[{Howard and Ruder(2018)}]{howard2018universal}
Jeremy Howard and Sebastian Ruder. 2018.
\newblock \href {https://doi.org/10.18653/v1/P18-1031} {Universal language model fine-tuning for text classification}.
\newblock In \emph{Proceedings of the 56th Annual Meeting of the Association for Computational Linguistics (Volume 1: Long Papers)}, pages 328--339, Melbourne, Australia. Association for Computational Linguistics.

\bibitem[{Hu et~al.(2022)Hu, Shen, Wallis, Allen-Zhu, Li, Wang, Wang, and Chen}]{hu2022lora}
Edward~J. Hu, Yelong Shen, Phillip Wallis, Zeyuan Allen-Zhu, Yuanzhi Li, Shean Wang, Lu~Wang, and Weizhu Chen. 2022.
\newblock \href {https://openreview.net/forum?id=nZeVKeeFYf9} {{L}o{RA}: Low-rank adaptation of large language models}.
\newblock In \emph{Proceedings of the Tenth International Conference on Learning Representations}, Virtual Event.

\bibitem[{Jeon et~al.(2025)Jeon, Kim, and Kim}]{jeon2025l4q}
Hyesung Jeon, Yulhwa Kim, and Jae-Joon Kim. 2025.
\newblock \href {https://doi.org/10.18653/v1/2025.acl-long.99} {L4q: Parameter efficient quantization-aware fine-tuning on large language models}.
\newblock In \emph{Proceedings of the 63rd Annual Meeting of the Association for Computational Linguistics (Volume 1: Long Papers)}, pages 2002--2024, Vienna, Austria. Association for Computational Linguistics.

\bibitem[{Kantharuban et~al.(2023)Kantharuban, Vuli{\'c}, and Korhonen}]{kantharuban2023quantifying}
Anjali Kantharuban, Ivan Vuli{\'c}, and Anna Korhonen. 2023.
\newblock \href {https://doi.org/10.18653/v1/2023.findings-emnlp.481} {Quantifying the dialect gap and its correlates across languages}.
\newblock In \emph{Findings of the Association for Computational Linguistics: EMNLP 2023}, pages 7226--7245, Singapore. Association for Computational Linguistics.

\bibitem[{Keller et~al.(2023)Keller, Sachdeva et~al.}]{leolm2023}
Bj{\"o}rn Keller, Nishant Sachdeva, and 1 others. 2023.
\newblock Leolm: Igniting german-language llm research.
\newblock \url{https://laion.ai/blog/leo-lm/}.
\newblock LAION blog post.

\bibitem[{Kirkpatrick et~al.(2017)Kirkpatrick, Pascanu, Rabinowitz, Veness, Desjardins, Rusu, Milan, Quan, Ramalho, Grabska-Barwinska, Hassabis, Clopath, Kumaran, and Hadsell}]{Kirkpatrick_2017}
James Kirkpatrick, Razvan Pascanu, Neil Rabinowitz, Joel Veness, Guillaume Desjardins, Andrei~A. Rusu, Kieran Milan, John Quan, Tiago Ramalho, Agnieszka Grabska-Barwinska, Demis Hassabis, Claudia Clopath, Dharshan Kumaran, and Raia Hadsell. 2017.
\newblock \href {https://doi.org/10.1073/pnas.1611835114} {Overcoming catastrophic forgetting in neural networks}.
\newblock \emph{Proceedings of the National Academy of Sciences}, 114(13):3521--3526.

\bibitem[{Koto et~al.(2025)Koto, Joshi, Mukhituly, Wang, Xie, Pal, Orel, Mullah, Turmakhan, Goloburda, Kamran, Ghosh, Jia, Mansurov, Togmanov, Banerjee, Laiyk, Sakip, Han, Kochmar, Aji, Singh, Jadhav, Katipomu, Kamboj, Choudhury, Gosal, Ramakrishnan, Mishra, Chandran, Sheinin, Vassilieva, Sengupta, and Nakov}]{koto2025sherkalachat}
Fajri Koto, Rituraj Joshi, Nurdaulet Mukhituly, Yuxia Wang, Zhuohan Xie, Rahul Pal, Daniil Orel, Parvez Mullah, Diana Turmakhan, Maiya Goloburda, Mohammed Kamran, Samujjwal Ghosh, Bokang Jia, Jonibek Mansurov, Mukhammed Togmanov, Debopriyo Banerjee, Nurkhan Laiyk, Akhmed Sakip, Xudong Han, and 15 others. 2025.
\newblock \href {https://openreview.net/forum?id=wRcTCcb0H5} {Sherkala-chat: Building a state-of-the-art llm for kazakh in a moderately resourced setting}.
\newblock In \emph{Proceedings of the Second Conference on Language Modeling (COLM)}.

\bibitem[{Le et~al.(2020)Le, Vial, Frej, Segonne, Coavoux, Lecouteux, Allauzen, Crabb{\'e}, Besacier, and Schwab}]{le-etal-2020-flaubert}
Hang Le, Lo{\"i}c Vial, Jibril Frej, Vincent Segonne, Maximin Coavoux, Benjamin Lecouteux, Alexandre Allauzen, Benoit Crabb{\'e}, Laurent Besacier, and Didier Schwab. 2020.
\newblock \href {https://aclanthology.org/2020.lrec-1.302/} {Flaubert: Unsupervised language model pre-training for french}.
\newblock In \emph{Proceedings of the 12th Language Resources and Evaluation Conference (LREC 2020)}, pages 2479--2490, Marseille, France. European Language Resources Association.

\bibitem[{Lee et~al.(2020)Lee, Yoon, Kim, Kim, Kim, So, and Kang}]{lee2019biobert}
Jinhyuk Lee, Wonjin Yoon, Sungdong Kim, Donghyeon Kim, Sunkyu Kim, Chan~Ho So, and Jaewoo Kang. 2020.
\newblock \href {https://doi.org/10.1093/bioinformatics/btz682} {{B}io{BERT}: a pre-trained biomedical language representation model for biomedical text mining}.
\newblock \emph{Bioinformatics}, 36(4):1234--1240.

\bibitem[{Lester et~al.(2021)Lester, Al-Rfou, and Constant}]{lester-etal-2021-power}
Brian Lester, Rami Al-Rfou, and Noah Constant. 2021.
\newblock \href {https://doi.org/10.18653/v1/2021.emnlp-main.243} {The power of scale for parameter-efficient prompt tuning}.
\newblock In \emph{Proceedings of the 2021 Conference on Empirical Methods in Natural Language Processing}, pages 3045--3059, Online and Punta Cana, Dominican Republic. Association for Computational Linguistics.

\bibitem[{Li and Liang(2021)}]{li2021prefixtuning}
Xiang~Lisa Li and Percy Liang. 2021.
\newblock \href {https://doi.org/10.18653/v1/2021.acl-long.353} {Prefix-tuning: Optimizing continuous prompts for generation}.
\newblock In \emph{Proceedings of the 59th Annual Meeting of the Association for Computational Linguistics and the 11th International Joint Conference on Natural Language Processing (Volume 1: Long Papers)}, pages 4582--4597, Online. Association for Computational Linguistics.

\bibitem[{Liu et~al.(2022)Liu, Tam, Muqeeth, Mohta, Huang, Bansal, and Raffel}]{liu2022fewshot}
Haokun Liu, Derek Tam, Mohammed Muqeeth, Jay Mohta, Tenghao Huang, Mohit Bansal, and Colin Raffel. 2022.
\newblock \href {https://proceedings.neurips.cc/paper_files/paper/2022/file/0cde695b83bd186c1fd456302888454c-Paper-Conference.pdf} {Few-shot parameter-efficient fine-tuning is better and cheaper than in-context learning}.
\newblock In \emph{Advances in Neural Information Processing Systems}, volume~35, pages 1950--1965.

\bibitem[{Lopez-Paz and Ranzato(2017)}]{lopezpaz2017gem}
David Lopez-Paz and Marc{'}Aurelio Ranzato. 2017.
\newblock \href {https://proceedings.neurips.cc/paper_files/paper/2017/file/f87522788a2be2d171666752f97ddebb-Paper.pdf} {Gradient episodic memory for continual learning}.
\newblock In \emph{Advances in Neural Information Processing Systems}, volume~30, pages 6467--6476.

\bibitem[{Mansha(2025)}]{mansha2025resourceefficient}
Imran Mansha. 2025.
\newblock \href {https://arxiv.org/abs/2510.05003} {Resource-efficient fine-tuning of llama-3.2-3b for medical chain-of-thought reasoning}.
\newblock \emph{Preprint}, arXiv:2510.05003.

\bibitem[{Martin et~al.(2020)Martin, Muller, Ortiz~Su{\'a}rez, Dupont, Romary, de~la Clergerie, Seddah, and Sagot}]{martin-etal-2020-camembert}
Louis Martin, Benjamin Muller, Pedro~Javier Ortiz~Su{\'a}rez, Yoann Dupont, Laurent Romary, {\'E}ric de~la Clergerie, Djam{\'e} Seddah, and Beno{\^i}t Sagot. 2020.
\newblock \href {https://doi.org/10.18653/v1/2020.acl-main.645} {{C}amem{BERT}: a tasty {F}rench language model}.
\newblock In \emph{Proceedings of the 58th Annual Meeting of the Association for Computational Linguistics}, pages 7203--7219, Online. Association for Computational Linguistics.

\bibitem[{Pfeiffer et~al.(2021)Pfeiffer, Kamath, R{\"u}ckl{\'e}, Cho, and Gurevych}]{pfeiffer-etal-2021-adapterfusion}
Jonas Pfeiffer, Aishwarya Kamath, Andreas R{\"u}ckl{\'e}, Kyunghyun Cho, and Iryna Gurevych. 2021.
\newblock \href {https://doi.org/10.18653/v1/2021.eacl-main.39} {{A}dapter{F}usion: Non-destructive task composition for transfer learning}.
\newblock In \emph{Proceedings of the 16th Conference of the European Chapter of the Association for Computational Linguistics: Main Volume}, pages 487--503, Online. Association for Computational Linguistics.

\bibitem[{{Proxecto NOS}(2024)}]{caraballo2024}
{Proxecto NOS}. 2024.
\newblock Llama-3.1-carballo: Galician language model.
\newblock \url{https://huggingface.co/proxectonos/Llama-3.1-Carballo-Instr1}.
\newblock Hugging Face model repository.

\bibitem[{Sarkar et~al.(2022)Sarkar, Lin, Sengupta, Lausen, Zha, and Mansour}]{sarkar2022parameter}
Soumajyoti Sarkar, Kaixiang Lin, Sailik Sengupta, Leonard Lausen, Sheng Zha, and Saab Mansour. 2022.
\newblock \href {https://neurips2022-enlsp.github.io/papers/paper_53.pdf} {Parameter and data efficient continual pre-training for robustness to dialectal variance in arabic}.
\newblock In \emph{Proceedings of the NeurIPS 2022 Workshop on Efficient Natural Language and Speech Processing (ENLSP)}, New Orleans, USA.

\bibitem[{{VietAI Research}(2023)}]{vinaLlama2023}
{VietAI Research}. 2023.
\newblock Vinallama: Vietnamese large language model.
\newblock \url{https://github.com/VietAI-Research/VinaLLaMA}.
\newblock GitHub repository.

\bibitem[{Yang et~al.(2019)Yang, Zhang, Tar, and Baldridge}]{yang2019pawsx}
Yinfei Yang, Yuan Zhang, Chris Tar, and Jason Baldridge. 2019.
\newblock \href {https://doi.org/10.18653/v1/D19-1382} {Paws-x: A cross-lingual adversarial dataset for paraphrase identification}.
\newblock In \emph{Proceedings of the 2019 Conference on Empirical Methods in Natural Language Processing and the 9th International Joint Conference on Natural Language Processing (EMNLP-IJCNLP 2019)}, pages 3687--3692, Hong Kong, China. Association for Computational Linguistics.

\end{thebibliography}

\appendix

\section{Training Configuration}
\label{app:train-config}

\noindent\textbf{Hardware.}
\begin{itemize}\setlength\itemsep{0pt}
  \item Devices ($d$): NVIDIA Tesla V100 (32\,GB) \emph{(single- or multi-GPU as noted below)}
  \item Precision: \texttt{fp16} with gradient checkpointing
\end{itemize}

\noindent\textbf{Batching and Effective Batch.}
Let $b$ be sequences per device (micro-batch), $a$ the gradient-accumulation steps, and $d$ the number of devices. The effective batch in \emph{sequences} is
\[
\text{EffBatch}_{\text{seq}} = b \times a \times d.
\]
With a sequence length of 1024 tokens, the effective batch in tokens is
$\text{EffBatch}_{\text{tok}} = \text{EffBatch}_{\text{seq}} \times 1024$.

\medskip
\noindent\textbf{Profiles used in our runs.}
\begin{itemize}\setlength\itemsep{2pt}
  \item \textbf{Profile A (1B CPT, single V100):} $b{=}4$, $a{=}8$, $d{=}1$ 
  $\Rightarrow$ $\text{EffBatch}_{\text{seq}}{=}32$ and $\text{EffBatch}_{\text{tok}}{=}32{,}768$.
  \item \textbf{Profile B (8B CPT, two V100s):} $b{=}1$, $a{=}8$, $d{=}2$
  $\Rightarrow$ $\text{EffBatch}_{\text{seq}}{=}16$ and $\text{EffBatch}_{\text{tok}}{=}16{,}384$.
\end{itemize}
(For larger models, we keep $a{=}8$ and adjust $b$ and $d$ to fit memory.)

\medskip
\noindent\textbf{Optimization.}
\begin{itemize}\setlength\itemsep{0pt}
  \item Objective: causal language modeling
  \item Optimizer: AdamW (weight decay $0.01$)
  \item LR schedule: cosine; base LR $1{\times}10^{-5}$; warmup ratio $0.1$
  \item Gradient clipping: $1.0$
  \item Checkpoint selection: best validation loss per epoch (no early stopping)
\end{itemize}

\noindent\textbf{LoRA (PEFT).}
\begin{itemize}\setlength\itemsep{0pt}
  \item Targets: attention ($qkvo$) and FFN ($\mathrm{up,gate,down}$)
  \item Rank $r{=}16$, $\alpha{=}32$, dropout $0.1$
\end{itemize}

\section{LoRA Hyperparameter Ablation}
\label{app:lora-ablation}

Our main experiments use LoRA with rank $r{=}16$, $\alpha{=}32$, and dropout $0.1$, following common practice. To validate these choices and provide guidance for future dialect adaptation work, we conducted a systematic ablation study on CroissantLLMChat-v0.1 using our Québec French corpus.

\subsection{Experimental Setup}

We evaluated four hyperparameter dimensions: (1) LoRA rank $r \in \{4, 8, 16, 32, 64\}$ with $\alpha = 2r$; (2) alpha-to-rank ratio $\alpha/r \in \{1, 2, 4\}$ with fixed $r{=}16$; (3) dropout $d \in \{0.0, 0.05, 0.1, 0.2\}$; and (4) target modules: $qv$ (query and value projections only), $qkvo$ (all attention projections), and $Full$ (attention + MLP layers). Each configuration was trained for 3 epochs with identical optimization settings. We report minimum validation loss and perplexity (PPL) achieved during training.

\subsection{Results}

Table~\ref{tab:lora-ablation} summarizes the ablation results. We highlight the best configuration within each ablation type.

\begin{table}[h]
\centering
\small
\setlength{\tabcolsep}{4pt}
\begin{tabular}{llrrcc}
\toprule
\textbf{Ablation} & \textbf{Config} & \textbf{$r$} & \textbf{$\alpha$} & \textbf{Loss} & \textbf{PPL} \\
\midrule
\multirow{5}{*}{Rank}
& $r=4$   & 4  & 8   & 3.057 & 21.27 \\
& $r=8$   & 8  & 16  & 3.021 & 20.52 \\
& $r=16$  & 16 & 32  & 2.980 & 19.68 \\
& $r=32$  & 32 & 64  & 2.930 & 18.74 \\
& $r=64$  & 64 & 128 & \textbf{2.870} & \textbf{17.63} \\
\midrule
\multirow{3}{*}{$\alpha/r$ $ratio$}
& $\alpha/r=1$ & 16 & 16 & 3.009 & 20.26 \\
& $\alpha/r=2$ & 16 & 32 & 2.980 & 19.68 \\
& $\alpha/r=4$ & 16 & 64 & \textbf{2.953} & \textbf{19.15} \\
\midrule
\multirow{4}{*}{Dropout}
& $d=0.00$ & 16 & 32 & \textbf{2.976} & \textbf{19.62} \\
& $d=0.05$ & 16 & 32 & 2.980 & 19.68 \\
& $d=0.10$ & 16 & 32 & 2.983 & 19.74 \\
& $d=0.20$ & 16 & 32 & 2.989 & 19.86 \\
\midrule
\multirow{3}{*}{Modules}
& $qv$ only    & 16 & 32 & 3.111 & 22.44 \\
& $qkvo$     & 16 & 32 & 3.072 & 21.59 \\
& $Full$ & 16 & 32 & \textbf{2.979} & \textbf{19.68} \\
\bottomrule
\end{tabular}
\caption{LoRA hyperparameter ablation on CroissantLLMChat-v0.1 with 3 epochs of CPT on Québec French. Best result per ablation type in \textbf{bold}.}
\label{tab:lora-ablation}
\end{table}

\subsection{Analysis}

\paragraph{Rank.} Higher rank consistently improves performance, with $r{=}64$ achieving $17\%$ lower perplexity than $r{=}4$ ($17.63$ vs.\ $21.27$). This suggests that dialect adaptation benefits from increased adapter capacity, likely because capturing lexical, morphological, and syntactic variations requires more expressive low-rank subspaces. However, trainable parameters scale linearly with rank. Examining the efficiency of each doubling: moving from $r{=}32$ to $r{=}64$ doubles the parameter count but yields only a $5.9\%$ additional perplexity reduction ($18.74$ to $17.63$), whereas $r{=}32$ already captures $70\%$ of the total improvement over $r{=}4$ (perplexity $18.74$ vs.\ $21.27$). For resource-constrained settings typical of low-resource dialect work, $r{=}32$ offers a favorable trade-off: near-optimal adaptation quality at half the parameter budget of $r{=}64$.

\paragraph{Alpha scaling.} The $\alpha/r$ $ratio$ controls the effective learning rate of the LoRA updates. We find that $\alpha/r{=}4$ slightly outperforms the standard $\alpha/r{=}2$ setting (loss $2.953$ vs.\ $2.980$), suggesting that dialect adaptation benefits from more aggressive updates to the low-rank matrices. This may reflect the need to provide stronger gradient updates to shift the model's lexical and syntactic distributions when adapting to a regional variety.

\paragraph{Dropout.} Contrary to the common practice of using dropout during fine-tuning, we find that \emph{zero dropout} yields the best results for dialect CPT (loss $2.976$ vs.\ $2.989$ at $d{=}0.2$). Performance degrades monotonically as dropout increases. We hypothesize that for continual pretraining on a focused domain corpus, the model benefits from fully absorbing dialect-specific vocabulary and expressions. Dropout may interfere with learning these low-frequency but important patterns.

\paragraph{Target modules.} Applying LoRA to all linear layers ($Full$) substantially outperforms attention-only configurations. This configuration achieves $12\%$ lower perplexity than $qv$ ($19.68$ vs.\ $22.44$), indicating that MLP layers play an important role in encoding dialect-specific lexical and semantic knowledge, not just the attention mechanisms. 

\subsection{Recommendations}

Based on these ablations, the optimal LoRA configuration for low-resource dialect CPT is: $r \geq 32$, $\alpha/r \geq 2$, dropout $= 0$, and full module targeting. Our main experiments used $r{=}16$, $\alpha{=}32$, and $d{=}0.1$ as a conservative baseline following prior work; the ablations suggest that more aggressive settings may yield further improvements. We leave exploration of these optimized settings across all models to future work.

\end{document}